%% file: main.tex
\documentclass{article}
\usepackage{ijcai26}

\usepackage[table]{xcolor} 

\usepackage{cite}

\usepackage{times}
\usepackage{soul}
\usepackage{url}
\usepackage[breaklinks,colorlinks,citecolor=teal,urlcolor=teal]{hyperref}
\usepackage[utf8]{inputenc}
\usepackage[small]{caption}
\usepackage{graphicx}
\usepackage{amsmath}
\usepackage{amsthm}
\usepackage{booktabs}
\usepackage{algorithm}
\usepackage{algorithmic}
\usepackage[switch]{lineno}
\definecolor{mygreen}{RGB}{85, 139, 47}
\definecolor{myblue}{RGB}{0, 0, 139}
\usepackage{cleveref}

\usepackage[edges]{forest}
\usepackage{xcolor}
\usetikzlibrary{arrows.meta,calc,positioning}

\definecolor{cTask}{HTML}{FFF2CC}
\definecolor{cPert}{HTML}{DAE8FC}
\definecolor{cRoute}{HTML}{E2F0D9}
\definecolor{cPipe}{HTML}{FCE4D6}
\definecolor{cDec}{HTML}{E6F2FF}
\definecolor{cRob}{HTML}{EDEDED}

\usepackage{tikz}
\usetikzlibrary{shapes,arrows,positioning,fit,calc,shadows}
\usepackage[edges]{forest} 

\usepackage{enumitem}
\usepackage{multirow}
\usepackage{rotating}
\usepackage{adjustbox}
\usepackage{array}
\usepackage{makecell}
\usepackage{pifont}
\usepackage{amssymb}
\usepackage{subcaption}
\newcommand{\cmark}{\ding{51}}
\definecolor{grpInfo}{HTML}{E3F2FD}
\definecolor{grpTask}{HTML}{E8F5E9}
\definecolor{grpEmbed}{HTML}{FFF3E0}
\definecolor{grpDecode}{HTML}{F3E5F5}
\definecolor{rowGray}{HTML}{F9F9F9}
\definecolor{color2d}{RGB}{56, 142, 60}
\definecolor{colordl}{RGB}{25, 118, 210}
\definecolor{color3d}{RGB}{230, 74, 25}
\definecolor{mytext}{RGB}{50, 50, 50}
\definecolor{mygray}{RGB}{160, 160, 160}
\definecolor{mylightgreen}{RGB}{120, 180, 120}
\definecolor{mylightblue}{RGB}{100, 160, 210}

\title{Intellectual Property Protection for 3D Gaussian Splatting Assets: A Survey}
\author{
Longjie Zhao$^{1}$
\and
Ziming Hong$^{1}$
\and
Jiaxin Huang$^{3}$
\and
Runnan Chen$^{1}$
\and
\\
Mingming Gong$^{2,3}$
\And
Tongliang Liu$^{1,3,*}$
\affiliations
$^{1}$Sydney AI Centre, The University of Sydney,\quad
$^{2}$The University of Melbourne\\
$^{3}$Mohamed bin Zayed University of Artificial Intelligence\\
\emails
lzha0538@uni.sydney.edu.au,\quad
hoongzm@gmail.com,\quad
jiaxin.huang@mbzuai.ac.ae,
mingming.gong@unimelb.edu.au,\quad
\{runnan.chen,tongliang.liu\}@sydney.edu.au
}

\begin{document}
\maketitle

\input{sec/0_abstract}

\input{sec/1_intro}

\input{sec/2_preliminary}

\input{sec/3_perturbations}

\input{sec/4_passive}

\input{sec/5_active}

\input{sec/6_robust}

\input{sec/7_future}

\input{sec/8_conclusion}

{
\small
\bibliographystyle{unsrt}
\bibliography{ref}

}

\end{document}

%% file: sec/0_abstract.tex
\begin{abstract}

3D Gaussian Splatting (3DGS) has become a mainstream representation for real-time 3D scene synthesis, enabling applications in virtual and augmented reality, robotics, and 3D content creation. Its rising commercial value and explicit parametric structure raise emerging intellectual property (IP) protection concerns, prompting a surge of research on 3DGS IP protection. However, current progress remains fragmented, lacking a unified view of the underlying mechanisms, protection paradigms, and robustness challenges. To address this gap, we present the first systematic survey on 3DGS IP protection and introduce a bottom-up framework that examines (i) underlying Gaussian-based perturbation mechanisms, (ii) passive and active protection paradigms, and (iii) robustness threats under emerging generative AI era, revealing gaps in technical foundations and robustness characterization and indicating opportunities for deeper investigation. Finally, we outline six research directions across robustness, efficiency, and protection paradigms, offering a roadmap toward reliable and trustworthy IP protection for 3DGS assets.
A collection of relevant papers is summarized and will be continuously updated at \url{https://github.com/tmllab/Awesome-3DGS-IP-Protection}.

\end{abstract}

%% file: sec/1_intro.tex
\begin{figure*}[t]
    \centering
    \includegraphics[width=\linewidth]{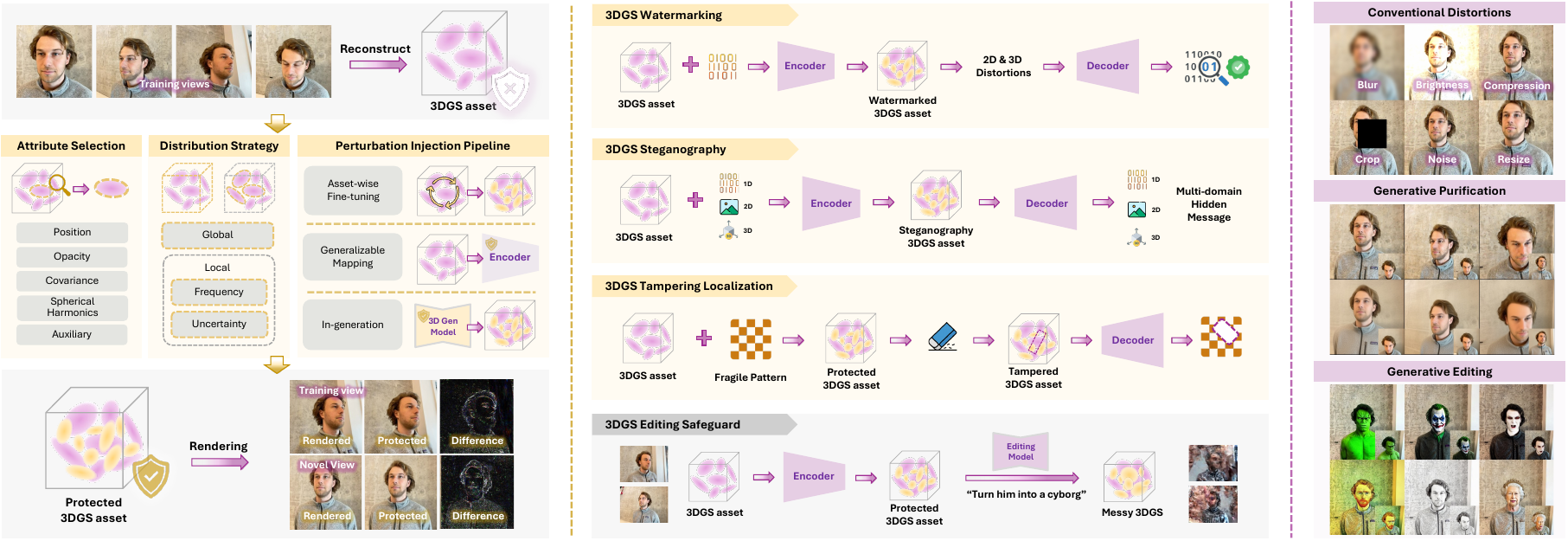}\\
    \small
     \hspace{5mm} (a) Gaussian-based perturbations \hspace{35mm} (b) 3DGS IP protection tasks \hspace{27mm} (c) Robustness \hspace{5mm}
    \\
    \vspace{-2mm}
    \caption{We introduce a \textit{bottom-up framework} that summarize 3DGS IP protection from (a) underlying mechanisms of Gaussian-based perturbation (\Cref{sec:perturbations}), through (b) 3DGS IP protection tasks (\Cref{sec:passive,sec:active}), to (c) emerging robustness threats in AIGC era (\Cref{sec:robustness}).}
    \label{fig:framework}
    \vspace{-4mm}
\end{figure*}

\vspace{-3mm}
\section{Introduction}

3D Gaussian Splatting (3DGS) \cite{kerbl20233d,sun2025intern,chen2024beyond} revolutionizes novel view synthesis by representing scenes or objects as collections of anisotropic 3D Gaussians with learnable attributes, offering superior rendering quality and real-time efficiency compared to existing 3D representations (e.g. Neural Radiance Fields (NeRF) \cite{mildenhall2021nerf}). These advantages have driven transformative applications across virtual/augmented reality \cite{fei20243d}, robotic perception \cite{tosi2024nerfs,chen2024ovgaussian,huang2025surprise3d}, AI-generated 3D content \cite{tang2024lgm,zou2024triplane}, and immersive gaming experiences \cite{bao20253d}. Its widespread adoption has created a burgeoning ecosystem of high-value digital assets.

However, the rapid adoption of 3DGS also introduces critical intellectual property (IP) protection challenges, where IP for digital 3D assets refers to the exclusive rights governing the ownership, control, and authorized use of creative or commercially valuable content (i.e., the legal entitlement to determine how such assets are accessed, modified, and distributed). Given both the high commercial value and the explicit parametric formulation of 3DGS assets, attackers are provided with high incentives and relative low technical barriers for IP misuse, including but not limited to: (i) unauthorized extraction, copying, and redistribution of asset parameters or rendered outputs \cite{huang2024gaussianmarker,yang2025concealgs}, and (ii) malicious generative editing that produces derivative assets while evading provenance or licensing \cite{zhao2025rdsplat,hong2025adlift}. Although existing IP protection techniques target 2D images or other 3D representations \cite{luo2023copyrnerf,song2024protecting,luo2025nerf,huang2024geometrysticker,wang2024unlearnable}, they cannot be directly applied to 3DGS: 2D protections typically vanish during 3DGS reconstruction, and 3DGS relies on Gaussian primitives that differ fundamentally from implicit NeRF or mesh-based representations. \textit{Accordingly, this highlights the urgent need for effective IP protection methods tailored to 3DGS.}

Researches on 3DGS IP protection has grown rapidly since 2024, with primary efforts focusing on \textit{3DGS watermarking} \cite{huang2024gaussianmarker,chen2025guardsplat} and \textit{3DGS steganography} \cite{zhang2024gs,wang2025stegagaussian}, and more recent studies extending to \textit{3DGS tampering localization} \cite{han2025gs} and \textit{3DGS editing safeguard} \cite{hong2025adlift}. Despite these progress, three critical gaps prevent the field from advancing systematically:
\begin{itemize}[leftmargin=*, topsep=0pt]\setlength{\parskip}{0pt}
    \item \textit{Fragmented research landscape.} 
    Existing works span multiple IP protection tasks, but lack a unified taxonomy to organize protection mechanisms, application scenarios, and their relationships. This prevents researchers from understanding the broader technical landscape and identifying optimal strategies under different security requirements.
    
    \item \textit{Lack of underlying technical analysis.}
    Across different 3DGS IP protection tasks, the techniques primarily rely on invisible 3D Gaussian-based perturbations.
    However, no prior work has systematically summarized the technical foundations or analyzed how they contribute to invisibility, protection capacity, and robustness.
    
    \item \textit{Incomplete robustness characterization.}
    While certain approaches assess robustness against conventional 2D and 3D distortions \cite{chen2025guardsplat,huang2024gaussianmarker}, the AI-Generated Content (AIGC) era \cite{lipman2022flow,li2025rethinking,chen2025semantics,lin2025beyond,fu2024discriminative,hou2024visual} introduces more sophisticated threats driven by the most advanced generative models, such as generative purification attacks \cite{huang2025can}, as well as the impact of generative editing on the integrity of protected assets \cite{haque2023instruct,hong2025adlift}, which are largely unexplored.
\end{itemize}

To bridge this gap, we present the first systematic survey on 3DGS IP protection. We introduce a \textit{bottom-up framework} (as shown in \Cref{fig:framework}) that progresses from (i) Gaussian-based perturbation, through (ii) 3DGS IP protection tasks, to (iii) emerging robustness threats in the AIGC era, and we categorize 24 existing methods under this framework in \Cref{tab:survey_3dgs_robustness}. Specifically, we first focus on underlying mechanisms of \textit{ Gaussian-based perturbation} (\Cref{sec:perturbations}), where we summarize existing techniques along three dimensions: attribute selection mechanism, distribution strategy, and perturbation injection pipelines. For \textit{3DGS IP protection tasks}, we distinguish existing methods between passive protection and active protection strategies depending on whether the goal is post-hoc tracing or proactive usability restriction, and provide a structured summary of both in \Cref{sec:passive} and \Cref{sec:active}, respectively. For \textit{emerging robustness threats} (\Cref{sec:robustness}), we begin with conventional 2D and 3D distortions and progressively expand to intentional generative purification attacks, and generative editing which can unintentional eliminate protective perturbations. Finally, in \Cref{sec:future}, we outline six future research directions and their associated challenges for 3DGS IP protection across three key dimensions: robustness, efficiency, and protection paradigms. We believe these directions will further advance robust 3DGS IP protection and support trustworthy deployment of 3DGS assets.

Our contributions are summarized as follows:
\begin{itemize}[leftmargin=*, topsep=0pt]\setlength{\parskip}{0pt}
    \item \textbf{Systematic survey.} We present the first systematic survey of 3DGS IP protection, structuring existing methods into passive and active protection paradigms.
    \item \textbf{Underlying technique analysis.} We analyze Gaussian-based perturbation, which serves as the primarily shared underlying technique across different 3DGS IP protection methods, spanning attribute selection mechanism, distribution strategy, and perturbation injection pipelines.
    \item \textbf{Robustness issues.} We highlight critical robustness risks, ranging from conventional 2D and 3D distortions to AIGC-driven generative purification and editing.
    \item \textbf{Future roadmap.} We outline six actionable research directions for securing 3DGS assets in the AIGC era.
\end{itemize}

%% file: sec/2_preliminary.tex
\vspace{-1mm}
\section{Preliminary}
\label{sec:preliminary}

\paragraph{3D Gaussian Splatting (3DGS).}
3DGS \cite{kerbl20233d,chen2024panoslam} represents a 3D scene or object as a collection of anisotropic 3D Gaussians $G = \{g_1, \dots, g_N\}$. Each Gaussian primitive $g_i$ is parameterized by a set of learnable attributes $\Theta_i = \{\mu_i, \Sigma_i, \alpha_i, \mathbf{c}_i\}$, where $\mu_i \in \mathbb{R}^3$ denotes the position, $\Sigma_i$ is the covariance matrix, $\alpha_i \in [0,1]$ is the opacity, and $\mathbf{c}_i$ represents the view-dependent color modeled by Spherical Harmonics (SH) coefficients. During rendering, these 3D Gaussians are projected into 2D image space. For a given pixel $p$, the accumulated color $\mathbf{C}(p)$ is computed via $\alpha$-blending of the $N$ sorted Gaussians overlapping with the pixel: $\mathbf{C}(p) = \sum_{i=1}^{N} c_i \alpha'_i \prod_{j=1}^{i-1} (1 - \alpha'_j),$ where $\alpha'_i$ is the effective opacity computed by multiplying $\alpha_i$ with the 2D Gaussian evaluation at pixel $p$. The differentiable nature of this rasterization pipeline allows for the back-propagation of gradients from 2D loss functions to 3D parameters $\Theta$, which forms the mathematical basis. Driven by its real-time rendering and high visual fidelity, 3DGS has rapidly become a dominant paradigm for 3D scene representation \cite{fei20243d,li2024urbangs,huang2025mllm,huang2025openinsgaussian}, thereby amplifying the need for IP protection mechanisms.

\paragraph{Taxonomy for 3DGS IP protection.} 

IP for digital 3D assets refers to the exclusive rights governing the ownership, control, and authorized use of creative or commercially valuable content (i.e., the legal entitlement to determine how such assets are accessed, modified, and distributed) \cite{kahng1998watermarking,li2025rethinking}. As shown in \Cref{fig:framework}(b), with the rise of 3DGS, corresponding IP protection techniques have introduced diverse objectives and mechanisms, making it necessary to organize them within a unified taxonomy. In this survey, we categorize existing methods based on their protective functionalities into two primary streams:

\begin{itemize}[leftmargin=*, topsep=0pt]\setlength{\parskip}{0pt}
\item \textit{Passive 3DGS IP protection} enables post-hoc verification and traceability of IP without actively restricting 3DGS asset functionality. These methods embed verifiable perturbations into the 3DGS asset to establish ownership or provenance, and to detect unauthorized tampering \cite{zhu2018hidden,qian2015deep,wu2019mantra,guo2023domain}. Existing passive approaches include: (i) \textit{3DGS watermarking} \cite{huang2024gaussianmarker}, which embeds robust identifiers for ownership verification; (ii) \textit{3DGS steganography} \cite{zhang2025securegs}, which conceals arbitrary payloads for covert communication or provenance; and (iii) \textit{3DGS tampering localization} \cite{han2025gs}, which reveals spatial regions undergoing malicious modification.

\item \textit{Active 3DGS IP protection} provides IP protection by proactively restricting the usability of 3DGS assets for unauthorized downstream tasks \cite{huang2021unlearnable,hong2025toward,wang2025catch,li2026webcloak}. Rather than tracing assets, these methods actively restrict unauthorized manipulation to enforce usage compliance and prevent misuse \cite{liang2023adversarial}. Existing approaches include \textit{3DGS editing safeguard} \cite{hong2025adlift}, which protects 3DGS assets against instruction-driven generative editing.

\end{itemize}

\paragraph{General 3DGS IP protection objective.} Regardless of the specific protection goals, we propose a general framework of 3DGS IP protection. Let $\mathcal{G}$ denote a dataset of original 3DGS assets, and $G\sim\mathcal{G}$ is a sampled asset.
We denote the protected version as $\tilde{G} = f_\theta(G)$, where $f_\theta$ denotes a protection operator parameterized by $\theta$ (e.g., learnable network weights or optimized Gaussian parameters). The general objective can be formulated as follows:

{
\setlength{\abovedisplayskip}{-7pt}
\setlength{\belowdisplayskip}{3pt}
\begin{align}
\max_\theta\ &
\underbrace{\smash{
\mathbb{E}_{G\sim\mathcal{G}}}
\big[
  s(f_\theta(G);\tau)
\big]
}_{\text{Protective capacity}\ \mathcal{L}_p(\theta)}
+\lambda
\underbrace{
\mathbb{E}_{G\sim\mathcal{G},\,T \sim \mathcal{T}_{\text{rob}}}
\big[
  s(T(f_\theta(G));\tau)
\big]
}_{\text{Robustness}\ \mathcal{L}_r(\theta)}
\nonumber
\\[-3pt]
\text{s.t.}\ &
\underbrace{
\mathbb{E}_{G\sim\mathcal{G},v \sim \mathcal{V}}
\big[
  d(R_v(G),R_v(f_\theta(G)))
\big]
}_{\text{Imperceptibility}\  \mathcal{C}_v(\theta)}\le\epsilon, 
\label{eq:genip}
\end{align}
}
which relies on three fundamental properties:
\begin{itemize}[leftmargin=*, topsep=0pt]\setlength{\parskip}{0pt}
    \item \textit{Imperceptibility} $\mathcal{C}_v(\theta)\le\epsilon$ 
    enforces that the protected asset $R_v(P_\theta(G))$ remains perceptually close to the original $R_v(G)$, where $d(\cdot,\cdot)$ measures perceptual distance and $\epsilon$ is a distortion budget. This ensures that IP protection does not degrade the photorealism or visual fidelity of 3DGS assets.

    \item \textit{Protective capacity} $\mathcal{L}_p(\theta)$ 
    evaluates the direct protection performance for a given task $\tau$ via a task-specific scoring function $s(\cdot;\tau)$, where higher scores correspond to stronger task compliance. Notably, different protection tasks may entail different operational objectives (\Cref{sec:passive} and \ref{sec:active}).

    \item \textit{Robustness} $\mathcal{L}_r(\theta)$
    characterizes how well the protection performance survives under transformations $T\sim\mathcal{T}_{\text{rob}}$, where $\mathcal{T}_{\text{rob}}$ may include conventional distortions as well as generative purification or editing, applied either to 2D rendered views or directly to the 3D asset space (\Cref{sec:robustness}). 

\end{itemize}

%% file: sec/3_perturbations.tex
\begin{table*}[!t]
\centering
\scriptsize

\newcolumntype{P}{>{\centering\arraybackslash}p{1.35em}}

\newcolumntype{C}{>{\centering\arraybackslash}p{1.35em}}

\definecolor{grpInfo}{HTML}{DAE8FC}
\definecolor{grpTask}{HTML}{FFF2CC}
\definecolor{grpDim}{HTML}{E2F0D9}
\definecolor{grpGauss}{HTML}{DAE8FC}
\definecolor{grpEmbed}{HTML}{E2EFDA}
\definecolor{grpDecode}{HTML}{E6E6E6}
\definecolor{grpGRP}{HTML}{E0E0E0}
\definecolor{grpAug}{HTML}{E6F5E6}
\definecolor{grp2D}{HTML}{FFF0E6}
\definecolor{grp3D}{HTML}{A02B93}
\setlength{\aboverulesep}{0pt}
\setlength{\belowrulesep}{0pt}
\setlength{\tabcolsep}{0.4pt}
\renewcommand{\arraystretch}{1.05}
\begin{tabular}{l c | C C C C C | C C C C | P P P | C C C | C C | C C | C C C C C C C C C C | C C C C C C C C C}
\toprule
\multicolumn{2}{c|}{\cellcolor{grpInfo!40}} 
& \multicolumn{12}{c|}{\cellcolor{grpGRP}\textbf{(a) Gaussian-based Perturbations}} 
& \multicolumn{5}{c|}{\cellcolor{grpTask}\textbf{(b) Task-specific}}
& \multicolumn{21}{c}{\cellcolor{grp3D!20}\textbf{(c) Robustness}} \\
\multicolumn{2}{c|}{\cellcolor{grpInfo!40}\multirow{-2}{*}{\textbf{Paper Information}}} 
& \multicolumn{5}{c|}{\cellcolor{grpGRP!30}\textbf{Attribute}} 
& \multicolumn{4}{c|}{\cellcolor{grpGRP!30}\textbf{Distribution}}
& \multicolumn{3}{c|}{\cellcolor{grpGRP!30}\textbf{Inject}} 
& \multicolumn{3}{c|}{\cellcolor{grpTask!30}\textbf{M-Dim}} 
& \multicolumn{2}{c|}{\cellcolor{grpTask!30}\textbf{P-Dim}} 
& \multicolumn{2}{c|}{\cellcolor{grp3D!10}\textbf{Aug}} 
& \multicolumn{10}{c|}{\cellcolor{grp3D!10}\textbf{2D Attacks}} 
& \multicolumn{9}{c}{\cellcolor{grp3D!10}\textbf{3D Attacks}} \\
\textbf{Method} & \textbf{Venue} 
& $\mu$ & $\Sigma$ & $\alpha$ & $c$ & $a$ 
& \rotatebox{90}{Glob} & \rotatebox{90}{Loc} & \rotatebox{90}{Freq} & \rotatebox{90}{Unce} 
& \rotatebox{90}{FT} & \rotatebox{90}{Map} & \rotatebox{90}{Gen} 
& \rotatebox{90}{1D} & \rotatebox{90}{2D} & \rotatebox{90}{3D} 
& \rotatebox{90}{2D} & \rotatebox{90}{3D} 
& \rotatebox{90}{2D} & \rotatebox{90}{3D} 
& \rotatebox{90}{JPEG} & \rotatebox{90}{Noise} & \rotatebox{90}{Blur} & \rotatebox{90}{Crop} & \rotatebox{90}{Rot} & \rotatebox{90}{Scale} & \rotatebox{90}{Photo} & \rotatebox{90}{Comb} & \rotatebox{90}{GenP} & \rotatebox{90}{GenE} 
& \rotatebox{90}{Prune} & \rotatebox{90}{Noise} & \rotatebox{90}{Rot} & \rotatebox{90}{Trans} & \rotatebox{90}{Scale} & \rotatebox{90}{Comp} & \rotatebox{90}{Clone} & \rotatebox{90}{GenP} & \rotatebox{90}{GenE} \\
\midrule
\multicolumn{38}{c}{\cellcolor{grpTask!10}\textbf{3DGS Watermarking}} \\
\hline

GaussianMarker~\cite{huang2024gaussianmarker} & NeurIPS'24 & \cmark & \cmark & \cmark & \cmark & & & \cmark & & \cmark& \cmark & & & \cmark & & & \cmark & \cmark & \cmark & \cmark & \cmark & \cmark & \cmark & \cmark& \cmark& \cmark & & \cmark&  & & \cmark & \cmark & \cmark & \cmark & & & & \\

\rowcolor{rowGray} WATER-GS~\cite{tan2024water} & arXiv'24 & \cmark & \cmark & \cmark & \cmark & & \cmark & & & & \cmark & & & \cmark & & & \cmark & & & \cmark & & & & & & & & & & & \cmark & \cmark & & & & &  & & \\

3D-GSW~\cite{jang20253d} & CVPR'25 & & \cmark & \cmark & \cmark & & \cmark& & \cmark & & \cmark & & & \cmark & & & \cmark & & & & \cmark & \cmark & \cmark & \cmark & \cmark & \cmark & & \cmark & & & \cmark & \cmark & & & & & \cmark & & \\

\rowcolor{rowGray} GuardSplat~\cite{chen2025guardsplat} & CVPR'25 & & & & \cmark & & \cmark & & & & \cmark & & & \cmark & & & \cmark & & \cmark & & \cmark & \cmark & \cmark & \cmark & \cmark & \cmark & \cmark & \cmark &  &\cmark & & & & & & & & & \\

MarkSplatter~\cite{huang2025marksplatter} & ACMMM'25 & & & & \cmark & & & \cmark & &\cmark & & \cmark & & \cmark & & & \cmark & \cmark & \cmark & \cmark & \cmark & \cmark & \cmark & \cmark & \cmark & \cmark & \cmark & & & & \cmark & \cmark & \cmark & \cmark & & & & & \\

\rowcolor{rowGray} MantleMark~\cite{luo2025mantlemark} & TIFS'25 & \cmark & \cmark & \cmark & \cmark & &  & \cmark & \cmark & & \cmark & & & \cmark & & & \cmark & & & & \cmark & \cmark & \cmark & \cmark & \cmark &  & \cmark & & & & & & & & & & & &  \\

NGS-Marker~\cite{anonymous2026ngsmarker} & OpenReview'25 & \cmark & \cmark & \cmark & \cmark & & & \cmark & & & \cmark&  & & \cmark & \cmark & & & \cmark & & \cmark & & & & & & & & & & & \cmark & \cmark & \cmark & \cmark & \cmark & & \cmark & & \cmark \\

Mark3DGS~\cite{xu2025markdgs} & OpenReview'25 & & & & \cmark & & & \cmark & \cmark & \cmark& \cmark & & & \cmark & & & & \cmark & \cmark & & \cmark & \cmark & & \cmark & \cmark & \cmark & & & & & \cmark & \cmark &  & &  &  & \cmark & &  \\

\rowcolor{rowGray} X-SG$^2$S~\cite{cheng2025x} & arXiv'25 & & & & \cmark & & & \cmark & & & &\cmark & & \cmark & \cmark & \cmark & & \cmark & &  & & & & & & & & & & & \cmark & & & & & & & &  \\

CompMarkGS~\cite{in2025compmarkgs} & arXiv'25 &\cmark &\cmark &\cmark &\cmark &\cmark & \cmark & & \cmark & & \cmark &  & & \cmark & & & \cmark & & \cmark & \cmark & \cmark & \cmark & \cmark & \cmark & \cmark & \cmark & & \cmark & & & \cmark & \cmark & & & & \cmark & \cmark & &  \\

\rowcolor{rowGray} GaussianSeal~\cite{li2025gaussianseal} & arXiv'25 &\cmark & \cmark& \cmark& \cmark& & \cmark & & \cmark & & & & \cmark & \cmark & & & \cmark & & & & \cmark & \cmark & \cmark & \cmark & \cmark & & \cmark & & & &\cmark & & & & & \cmark & & &  \\

GS-Marker~\cite{li2025gs} & arXiv'25 & \cmark & \cmark & \cmark & \cmark & & & \cmark & & & & \cmark & & \cmark & & & \cmark & & \cmark & \cmark & \cmark & \cmark & & \cmark & \cmark & & & & & & \cmark & & & & & & & & \\

\rowcolor{rowGray} RDSplat~\cite{zhao2025rdsplat} & arXiv'25 & \cmark & \cmark & \cmark & \cmark & & & \cmark & \cmark & & \cmark & & & \cmark & & & \cmark & & \cmark & & \cmark & \cmark & \cmark & \cmark & \cmark & \cmark & \cmark &&  & \cmark & & & & & & & & & \\

\hline
\multicolumn{38}{c}{\cellcolor{grpTask!10}\textbf{3DGS Steganography}}\\
\hline

\rowcolor{rowGray} GS-Hider~\cite{zhang2024gs} & NeurIPS'24 & & & & \cmark &\cmark & \cmark & & & & \cmark &  & & & \cmark & \cmark & & \cmark & & & & & & & & & & & &  & \cmark & & & & & & & & \\

StegaGaussian~\cite{wang2025stegagaussian} & KBS'25 & & & & \cmark & & \cmark & & \cmark & & \cmark &  & & & \cmark & & \cmark & & \cmark & & \cmark & \cmark & \cmark & \cmark & \cmark & \cmark & &  & & & \cmark & \cmark & \cmark & \cmark & & & & & \\

\rowcolor{rowGray} KeySS~\cite{ren2025all} & arXiv'25 & \cmark & \cmark & \cmark & \cmark & & \cmark & & & & \cmark &  & & & & \cmark & & \cmark & & & & & & & & & & & &  & & & & & & & & & \\

Hide A Bit~\cite{sasaki2025hide} & SIGGRAPH'25 & \cmark & \cmark & \cmark & \cmark & & \cmark & & & &--& -- &-- & & & \cmark & & \cmark & & & & & & & & & & & &  & & & & & & & & & \\

\rowcolor{rowGray} InstantSplamp~\cite{li2025instantsplamp} & ICLR'25 &\cmark &\cmark & \cmark& \cmark& & \cmark & & & & & & \cmark & \cmark & \cmark & & \cmark & & & & \cmark & \cmark & \cmark & & &  & & & & & & & & & & & & & \\

SecureGS~\cite{zhang2025securegs} & ICLR'25 & \cmark & \cmark & \cmark & \cmark &\cmark & & \cmark & & & \cmark &  & & \cmark & \cmark & \cmark & \cmark & \cmark& & & &  & & & & & & &  & & \cmark & \cmark & & & & &  & & \\

\rowcolor{rowGray} ConcealGS~\cite{yang2025concealgs} & ICASSP'25 & \cmark & \cmark & \cmark & \cmark & & \cmark & & & & \cmark &  & & & \cmark & & \cmark & &  & & \cmark &  & \cmark &   && & & & & & & & & & & & & & \\

Splats in Splats~\cite{guo2024splats} & AAAI'26 & & & \cmark & \cmark & & \cmark & &  & & \cmark &  & & &  & \cmark & & \cmark & & & & & & & & & & & &  & \cmark & \cmark & & & & & & & \\

\hline
\multicolumn{38}{c}{\cellcolor{grpTask!10}\textbf{3DGS Tampering Localization}}\\
\hline

\rowcolor{rowGray} GS-Checker~\cite{han2025gs} & AAAI'26 &  &  &  &  & \cmark &  & \cmark & & & \cmark & & & -- & -- & -- & \cmark & \cmark & & & & \cmark & \cmark & & & & & & & &  & \cmark & & & \cmark & & & & \cmark \\

\hline
\multicolumn{38}{c}{\cellcolor{grpTask!10}\textbf{3DGS Editing Safeguard}}\\
\hline

DEGauss~\cite{meng2025degauss} & NeurIPS'25 & \cmark & \cmark & \cmark & \cmark & & \cmark & & & & \cmark & & & -- & -- & -- &  & \cmark & \cmark & & & & & & & &  & & & & & & & & & & & & \cmark \\

\rowcolor{rowGray} AdLift~\cite{hong2025adlift} & arXiv'25 & \cmark & \cmark & \cmark & \cmark & & \cmark & & & & \cmark & & & -- & -- & -- & \cmark & \cmark & \cmark & & \cmark & & & & & & & & \cmark & \cmark & & & & & & & & & \cmark \\

\bottomrule
\end{tabular}

\vspace{-1mm}

\begin{flushleft}
    \scriptsize
    \hangindent=1.5em
    \hangafter=1
    \textbf{(a)} 
    \textbf{Attribute} -- $\mu$: position; $\Sigma$: covariance; $\alpha$: opacity; $c$: SH coefficients; $a$: auxiliary attributes. 
    \textbf{Distribution} -- 
    Glob: perturbations embedded into all Gaussians;
    Loc: perturbations embedded into a subset of Gaussians;
    Freq: perturbations embed via fourier frequency domain;
    Unce: uncertainty-guided embedding.
    \textbf{Inject} -- 
    FT: asset-wise fine-tuning;
    Map: generalizable mapping;
    Gen: in-generation injection.
    \\        
    \textbf{(b)} 
    \textbf{M-Dim} (embedded message dimension) -- 1D: bit sequence/text, 2D: image, 3D: 3D data;
    \textbf{P-Dim} (protective dimension): whether the task-specific protection can be conduct on 2D or 3D.
    \\
    \textbf{(c)}
    \textbf{Aug}: robustness augmentations applied in the 2D or 3D domain in training.
    \textbf{2D Attacks} -- 
    JPEG: jpeg compression; 
    Noise: Gaussian noise; 
    Blur: Gaussian blur;
    Crop; Rot: rotation; Scale;
    Photo: photometric (brightness, contrast);
    Comb: combined distortions; 
    GenP: generative purification;
    GenE: generative editing. 
    \textbf{3D Attacks} -- 
    Prune: sparsification, dropout, crop; 
    Noise; 
    Rot: rotation; 
    Trans: translation; 
    Scale;
    Comp: model compression/quantization;
    Clone: asset cloning;
    GenP: generative purification;
    GenE: generative editing.\\
    \cmark~indicates the method satisfies the corresponding criterion; blank indicates it does not; -- indicates the criterion is not applicable to this method.
\end{flushleft}
\vspace{-4mm}

\caption{Summary of 24 existing 3DGS IP protection methods categorized by task type and characterized by: (a) Gaussian-based perturbation strategies, (b) task-specific protective capabilities, and (c) robustness.}
\label{tab:survey_3dgs_robustness}
\vspace{-2mm}
\end{table*}

\vspace{-2mm}
\section{Gaussian-based Perturbations}
\label{sec:perturbations}

A underlying technique shared between 3DGS IP protection methods is to embed Gaussian-based perturbations, as shown in \Cref{fig:framework}(a). Different perturbation strategies inherently shape the trade-off among protection capacity, visual fidelity, and robustness. For clarity, we categorize Gaussian-based perturbations along three dimensions: (i) attribute selection mechanism, (ii) distribution strategy, and (iii) injection pipelines.

\vspace{-1mm}
\subsection{Attribute Selection Mechanism}

The most fundamental strategy lies in deciding which Gaussian attributes are permitted to carry perturbations. Gaussian parameters are inherently heterogeneous because different attributes exhibit distinct perceptual sensitivity, embedding capacity, and robustness, resulting in different trade-offs when chosen as perturbation carriers. Existing strategies can be categorized into SH-only perturbation, hybrid-attributes perturbation, and auxiliary attribute coupling.

\paragraph{Spherical Harmonics perturbation.}
SH coefficients are a popular choice for perturbation embedding in 3DGS \cite{huang2025marksplatter,chen2025guardsplat,cheng2025x,zhang2024gs,wang2025stegagaussian}, as modifying SH only affects color/appearance while preserving the underlying Gaussian geometry (e.g., position and covariance), thus maintaining high visual fidelity. 
To avoid obvious color distortions \cite{ren2025all}, existing methods usually introduce SH offsets as residual signals \cite{chen2025guardsplat}, partition higher-order SH bands \cite{cheng2025x}, or replace SH with coupled latent features \cite{zhang2024gs}. 
However, the high-dimensional and complex structure of SH coefficients leads to unstable optimization \cite{ren2025all}. Moreover, recent studies report that SH perturbation suffers from high statistical detectability \cite{ren2025all} and even insufficient protective capability against editing \cite{hong2025adlift}.

\paragraph{Hybrid-attributes perturbation.}

More methods adopt hybrid-attribute perturbation, where perturbations are injected not only into SH coefficients but also into geometry-related Gaussian attributes (e.g., position, covariance, opacity) \cite{hong2025adlift,meng2025degauss}.
This strategy enhances protection capability by reducing reliance on single attribute. However, geometry attributes are highly sensitive to perturbations, e.g., even minor deviations in \textit{position} can significantly degrade rendering quality \cite{chen2025guardsplat,li2025gs}, while perturbing covariance parameters often introduces noticeable artifacts, particularly near object boundaries \cite{chen2025guardsplat,huang2025marksplatter}.
As a result, although hybrid-attribute perturbation can strengthen protection and avoid isolated color distortion, it may also incur broader visual degradation due to geometric sensitivity, and typically requires more complex optimization to balance visual quality and protective capability on challenging protection tasks \cite{hong2025adlift,meng2025degauss}.

\paragraph{Auxiliary attribute coupling. } 

A small number of methods attach auxiliary protection attributes to the original Gaussians, i.e., they externally append new attributes dedicated to carrying protective perturbations, rather than modifying any native Gaussian attributes \cite{zhang2024gs,zhang2025securegs}.
Since these auxiliary channels do not interfere with the original parameters, they can fully preserve visual quality and support high-capacity embedding. However, since these auxiliary channels are not coupled with semantically meaningful attributes or rendering-critical parameters, adversaries may prune or sanitize them without degrading visual quality, thus limiting their practical robustness.

\vspace{-1mm}
\subsection{Distribution Strategy}

The \textit{distribution strategy} determines how perturbations are distributed across the Gaussian primitives of a 3DGS asset. Several methods selectively perturb a subset of Gaussians with specific distributions (referred to as \textit{local strategies}), typically using \textit{Fourier frequency} or \textit{uncertainty} as guidance for subset selection. In contrast, methods that do not impose any explicit selection constraints on Gaussian primitives, and instead apply perturbations across the entire asset, are referred to as a \textit{global strategy}.

\paragraph{Fourier frequency guidance.}

Recent methods observe that different frequency domains exhibit distinct robustness and fidelity trade-offs, assigning high frequency to visual fidelity and low frequency to robustness~\cite{wang2024gaussianeditor}. 
In general, high-frequency components are closely tied to fine-grained appearance details and are therefore more perceptually sensitive, while low-frequency components encode more stable structural and semantic information \cite{li2025gaussianseal}. Perturbations embedded in high-frequency domains are often fragile, as they are suppressed by common compression pipelines~\cite{jang20253d,in2025compmarkgs} and disrupted by reconstruction or generative editing processes \cite{lu2024robust,zhao2025rdsplat}.
As a result, purely high-frequency embedding is typically avoided in robustness-oriented protection. In contrast, several methods prioritize low-frequency embedding to enhance survivability under distortions \cite{lu2024robust,zhao2025rdsplat}.
To balance the complementary properties of different frequency domains, hybrid-frequency local strategies have also been explored \cite{jang20253d,in2025compmarkgs}.
This hybrid design demonstrates that coordinating perturbations across multiple frequency domains can achieve a more favorable balance between imperceptibility and robustness than single-domain strategies.

\paragraph{Uncertainty guidance.} 
In 3DGS, uncertainty reflects the sensitivity of rendering quality to perturbations in individual Gaussian primitives.
GaussianMarker~\cite{huang2024gaussianmarker} and MarkSplatter~\cite{huang2025marksplatter} exploit this property by embedding watermarks into Gaussians with high uncertainty, which contribute less to the final rendered appearance, such as those near object boundaries. Since perturbations applied to these Gaussians induce minimal perceptual changes, such strategies enable imperceptible modifications while maintaining reliable message extraction.
However, because the number and spatial distribution of high-uncertainty Gaussians are inherently limited, such strategies often face constraints in achievable embedding capacity and flexibility.

\paragraph{Global strategy.} 
The global strategy distributes perturbations uniformly across all Gaussian primitives without explicit subset selection.
This design is widely adopted in existing methods~\cite{ren2025all,han2025gs,meng2025degauss,hong2025adlift}, particularly in steganographic settings \cite{ren2025all,wang2025stegagaussian} where high embedding capacity and strong protection ability are required \cite{hong2025adlift}.
By leveraging the entire Gaussian representation, global strategies maximize information capacity and avoid dependence on specific spatial or spectral regions.
However, because perturbations are applied broadly, this strategy typically requires careful optimization to preserve visual fidelity and prevent perceptible degradation \cite{hong2025adlift,meng2025degauss}.

\subsection{Perturbation Injection Pipeline}

Existing 3DGS IP protection methods mainly adopt three pipelines for perturbation injection, including \textit{asset-wise fine-tuning}, \textit{generalizable mapping}, and \textit{in-generation injection}. 

\paragraph{Asset-wise fine-tuning.} The basic but broadly adopted strategy for per-asset protection is asset-wise fine-tuning.
This strategy is flexible because it can directly optimize Gaussian parameters with customizable objectives and distortion models, enabling scene-specific robustness-fidelity trade-offs \cite{hong2025adlift,han2025gs,chen2025guardsplat}.
However, the required per-asset optimization limits generalization to new assets \cite{huang2025marksplatter} and incurs time costs that scale linearly with the number of assets.

\paragraph{Generalizable mapping.} 
Generalizable feed-forward mapping trains a universal encoder on large-scale 3DGS datasets, thus enabling perturbation injection into arbitrary novel 3DGS assets in a single forward pass \cite{li2025gs,huang2025marksplatter,cheng2025x}. Several recent 3DGS watermarking methods adopt this strategy \cite{huang2025marksplatter,li2025gs}. However, complex perturbation modeling introduces training instability \cite{li2025gs}, and joint encoder-decoder training on large-scale 3D datasets incurs substantial multi-GPU overhead \cite{huang2025marksplatter,li2025gs}, leading to high initial deployment costs.

\paragraph{In-generation injection.} 
In-generation injection integrates perturbations directly into 3DGS generative models \cite{xiang2025structured,li2025gaussianseal,li2025instantsplamp}, enabling assets to be born with protective perturbations at creation time. This pipeline achieves high efficiency for large-scale content production \cite{li2025instantsplamp} and strong robustness, as perturbations are deeply embedded within the generative pipeline, either via cross-attention on intermediate features \cite{li2025instantsplamp} or through U-Net block outputs \cite{li2025gaussianseal}. However, such approaches can only protect assets generated by the corresponding 3DGS generative model and are not applicable to pre-existing content.

%% file: sec/4_passive.tex
\vspace{-1mm}
\section{Passive 3DGS IP Protection}
\label{sec:passive}

Passive 3DGS IP protection enables post-hoc verification and traceability without actively restricting 3DGS asset functionality. In this section, we summarize three passive protection paradigms: 3DGS watermarking, 3DGS steganography, and 3DGS tampering localization.

\vspace{-1mm}
\subsection{3DGS Watermarking}
\label{subsec:watermarking}

As 3DGS assets becomes increasingly circulated, reused, and redistributed across creative and commercial workflows, ensuring provable ownership and traceable usage of assets has become a critical requirement. 3DGS Watermarking fulfills this need by embedding verifiable identifiers that enable reliable copyright attribution even after assets undergo transformations or distribution through untrusted channels. 

The task of 3DGS watermarking is to embed a binary message $\mathbf{m} \in \mathcal{M} \subset \{0,1\}^L$ sampled from a length-$L$ space $\mathcal{M}$ into an original 3DGS asset $G$ to obtain a protected (watermarked) asset $f_\theta(G)$. As such, the task-specific scoring function $s(\cdot;\tau)$ in \Cref{eq:genip} can be instantiated as the log-likelihood of correctly decoding $\mathbf{m}$ from the protected asset:

{
\setlength{\abovedisplayskip}{-7pt}
\setlength{\belowdisplayskip}{3pt}
\begin{align}
    s(f_\theta(G);\tau)
    = \log P\big(\mathbf{m}\big|D\big(\Phi(f_\theta(G))\big)\big),
\end{align}
}where $D$ denotes a watermark decoder and $\Phi$ is a 2D/3D feature interface that maps the (possibly rendered) protected asset $f_\theta(G)$ into the input space of $D$.

As shown in \Cref{tab:survey_3dgs_robustness}, existing 3DGS watermarking methods embed low-capacity binary identifiers into Gaussian representations to enable ownership verification under distortions, typically by coupling watermark objectives with 3DGS optimization. GaussianMarker~\cite{huang2024gaussianmarker} introduces uncertainty-aware embedding by modulating perturbations using gradient-based uncertainty, jointly perturbing positions and SH coefficients to resist noise and pruning.
GuardSplat~\cite{chen2025guardsplat} improves robustness via adversarial training with simulated compression and noise, focusing on high-contribution Gaussians.
RDSplat~\cite{zhao2025rdsplat} targets diffusion-based generative editing by embedding watermarks into low-frequency covariance components.
3D-GSW~\cite{jang20253d} leverages frequency-guided densification and adaptive attribute weighting, while NGS-Marker~\cite{anonymous2026ngsmarker} embeds watermarks into local primitives to support partial infringement protection.
CompMarkGS~\cite{in2025compmarkgs} emphasizes compression robustness through quantization-aware training on anchor features. To improve efficiency and generalization, GS-Marker~\cite{li2025gs} adopts a dual-network design enabling cross-scene watermarking, MarkSplatter~\cite{huang2025marksplatter} reformulates Gaussians into grid-based splatter images for CNN-based embedding, and WATER-GS~\cite{tan2024water} supports both white-box and black-box extraction via view-agnostic optimization.
Beyond direct embedding, MantleMark~\cite{luo2025mantlemark} migrates watermarks from multi-view images during reconstruction, while GaussianSeal~\cite{li2025gaussianseal} integrates watermarking into generative 3DGS pipelines to produce born-secure assets.

\vspace{-1mm}
\subsection{3DGS Steganography}
\label{subsec:steganography}

Certain scenarios require covert and non-disruptive embedding of auxiliary information (e.g., licensing metadata, user identifiers, or scene annotations) for purposes such as secure communication or provenance binding. Steganography fulfills this need by enabling high-capacity and imperceptible embedding of multimodal payloads into 3DGS assets.

The task of 3DGS steganography is to embed a secret payload $\mathbf{p} \in \mathcal{P}$ sampled from a payload space $\mathcal{P}$ (e.g., a bit sequence, a 2D image, or a 3D Gaussian/volumetric asset) into a cover 3DGS asset $G$ to obtain a stego counterpart $f_\theta(G,\mathbf{p})$. We instantiate the task-specific scoring function $s(\cdot;\tau)$ in \Cref{eq:genip} as the negative payload reconstruction loss:

{
\setlength{\abovedisplayskip}{-7pt}
\setlength{\belowdisplayskip}{3pt}
\begin{align}
    s(f_\theta(G,\mathbf{p});\tau)
    = -\mathcal{L}_{\text{msg}}\big(\mathbf{p},\,\mathcal{D}(\Phi(f_\theta(G,\mathbf{p})))\big),
\end{align}
}where $\mathcal{D}$ denotes a stego payload decoder and $\Phi$ is a 2D/3D feature interface that maps the (possibly rendered) stego asset $f_\theta(G,\mathbf{p})$ into the input space of $\mathcal{D}$.

3DGS steganography aims to maximize payload capacity while preserving high visual fidelity, as summarized in \Cref{tab:survey_3dgs_robustness}.
Unlike watermarking, steganographic methods prioritize imperceptibility and information throughput, with limited emphasis on robustness against deliberate distortions. Most approaches embed messages by directly modulating appearance-related Gaussian attributes. GS-Hider~\cite{zhang2024gs} and ConcealGS~\cite{yang2025concealgs} achieve high-capacity embedding by modulating SH coefficients or color attributes with minimal visual degradation.
Splats in Splats~\cite{guo2024splats} further increases capacity by embedding 3D content via bit-shifting and opacity mapping, preserving structural integrity.
GaussianStego~\cite{li2024gaussianstego} introduces a generalizable encoder-decoder pipeline for cross-scene steganography.
Hide A Bit~\cite{sasaki2025hide} enables ultra-fast embedding through LSB manipulation combined with RSA encryption, without optimization.
KeySS~\cite{ren2025all} enforces access control using a key-secured decoder and a 3D-Sinkhorn metric, while SecureGS~\cite{zhang2025securegs} adopts a decoupled encryption scheme within Scaffold-GS anchors to balance security and fidelity.
StegaGaussian~\cite{wang2025stegagaussian} hides messages in high-frequency domains via frequency decomposition, whereas X-SG$^2$~\cite{cheng2025x} exploits cross-modal redundancy.
For dynamic scenes, Hide-in-Motion~\cite{liu2025hide} embeds payloads into spatiotemporal deformation fields of 4DGS.
Finally, InstantSplamp~\cite{li2025instantsplamp} performs in-generation steganography by injecting messages through cross-attention layers during generative inference.

\vspace{-1mm}
\subsection{3DGS Tampering Localization}
\label{subsec:tampering}

As 3DGS assets undergo editing, compositing, and transformation across production pipelines, unauthorized or malicious modifications may compromise authenticity and create trustworthiness concerns. 3DGS tampering localization focuses on detecting and spatially identifying altered regions to support integrity verification and forensic analysis. 

GS-Checker~\cite{han2025gs} is the first and currently the only 3DGS tampering localization method. It does not require pre-embedded perturbations; instead, it directly analyzes tampered 3DGS assets. By attaching a 3D tampering attribute to each Gaussian and employing a 3D contrastive mechanism to reveal local inconsistencies, GS-Checker detects suspect manipulations directly in the 3D domain.

%% file: sec/5_active.tex
\vspace{-1mm}
\section{Active 3DGS IP Protection}
\label{sec:active}

Active 3DGS IP protection proactively limits unauthorized downstream use of 3DGS assets \cite{li2025rethinking}. Current active methods predominantly target editing safeguards.

\vspace{-1mm}
\subsection{3DGS Editing Safeguard}
 
Recent 3DGS editing works
enable faithful manipulation of 3DGS assets and advance 3D content creation \cite{wu2024gaussctrl,lee2025editsplat,chen2024dge,wang2024view,chen2024gaussianeditor}. However, they also expose 3DGS assets to significant risks of unauthorized or malicious editing, potentially leading to identity deception, misinformation, or reputational damage \cite{liang2023adversarial,xue2024toward,shan2023glaze}. This highlights the urgency of developing effective 3DGS editing safeguard methods.

The task of 3DGS editing safeguard to learn a safeguard operator $f_\theta$ that produces a protected asset $f_\theta(G)$ for an original 3DGS asset $G$ such that instruction-driven editing either induces \textit{minimal semantic change} or \textit{drastically degrades editing quality} \cite{liang2023mist}. 
Thus, the task-specific scoring function $s(\cdot;\tau)$ in \Cref{eq:genip} can be written as the negative expected editing success under an editing model $\mathcal{E}_\phi$:

{
\setlength{\abovedisplayskip}{-7pt}
\setlength{\belowdisplayskip}{3pt}
\begin{align}
    s(f_\theta(G);\tau)
    = -\,\mathbb{E}_{\mathbf{e}}\!\left[
        \mathcal{L}_{\text{edit}}\!\left(
            \mathcal{E}_\phi\!\left(\Phi(f_\theta(G)),\mathbf{e}\right),
            \Phi(G)
        \right)
    \right],
    \label{eq:3dgseditguard}
\end{align}
}where $\mathbf{e}$ is an editing instruction sampled from an instruction space $\mathcal{E}$, $\Phi$ is a 2D/3D feature interface into the editing model, and $\mathcal{L}_{\text{edit}}$ measures editing success (e.g., semantic deviation or instruction alignment).

\Cref{eq:3dgseditguard} can be seen as an adversarial attack for the editing model \cite{liang2023adversarial,xue2024toward,liang2023mist}.
AdLift \cite{hong2025adlift} prevents instruction-driven editing by lifting strictly bounded 2D adversarial perturbations into a 3D Gaussian safeguard. It optimizes these safeguard Gaussians via a tailored lifted projected gradient descent (PGD) \cite{madry2017towards,huang2022harnessing,lin2023eliminating,lin2023over,lin2024layer} that truncates gradients from the editing model at the rendered image level and applies projected gradients to enforce strictly image-space invisable constraints. The final perturbations are propagated to Gaussians through an image-to-Gaussian fitting stage.
DEGauss \cite{meng2025degauss} improves cross-view robustness through a view-focal gradient fusion module that prioritizes gradients from challenging viewpoints. It further enhances the adversarial objective via dual discrepancy optimization, jointly maximizing semantic deviation and directional bias of the guidance signal to better disrupt editing trajectories.

%% file: sec/6_robust.tex
\vspace{-1mm}
\section{Robustness Issues}
\label{sec:robustness}

Robustness is essential for 3DGS IP protection, as adversaries may attempt to weaken or erase protection through conventional distortions or generative-based methods, affecting either 2D renders or the 3DGS space. As shown in \Cref{fig:framework}(c), we summarize three robustness challenges considered in existing 3DGS IP protection: (i) conventional 2D and 3D distortions, (ii) generative purification, and (iii) generative editing.

\paragraph{Conventional 2D and 3D distortions.} Since perturbations embedded in 3DGS are ultimately projected onto 2D rendered images, \textit{conventional 2D distortions} \cite{huang2023robust} can effectively disrupt such perturbations. These include geometric transformations (e.g., rotation and cropping \cite{jang20253d}), photometric variations (e.g., brightness and contrast adjustment \cite{huang2025marksplatter}), and signal degradations such as Gaussian blur, noise injection, and JPEG compression \cite{huang2024gaussianmarker}. 
Moreover, \textit{3D distortions} include geometric distortions and degradation distortions.
Geometric distortions explicitly alter the spatial configuration of Gaussian primitives. Pruning, cloning, and densification modify Gaussian density and distribution \cite{zhang2024gs,guo2024splats}, diluting protective perturbations. Spatial transformations including rotation, scaling, and translation perturb Gaussian attributes like positions and covariances, thereby misaligning geometry-dependent encodings \cite{huang2024gaussianmarker,anonymous2026ngsmarker}.  
In contrast, degradation distortions impair fidelity: Gaussian noise corrupts geometry and appearance jointly \cite{jang20253d,huang2025marksplatter}, while model compression suppresses low-amplitude perturbations through quantization and parameter merging \cite{zhang2025securegs}.

\paragraph{Generative purification.} Generative purification methods \cite{cao2023impress,zhao2024can} aim to map inputs back to their natural data distribution, thereby \textit{intentionally} eliminating protective perturbations while preserving fidelity. In the 2D domain, generative purification methods such as DiffPure \cite{nie2022diffusion}, Impress \cite{cao2023impress}, GrIDPure \cite{zhao2024can}, and PDM-Pure \cite{xue2024pixel} leverage diffusion-based reverse processes to systematically eliminate adversarial noise. Besides, 3D-specific methods focus on intrinsic 3DGS structures. Approaches like GSPure \cite{huang2025can} and GMEA \cite{zeng2025fading} exploit the redundancy of Gaussians, pruning elements that are discriminative for decoding yet contribute minimally to reconstruction. These methods effectively compromise mainstream schemes \cite{huang2024gaussianmarker,zhang2024gs,guo2024splats,zhang2025securegs} by balancing visual fidelity with attack efficacy, highlighting generative purification as a critical yet underexplored vulnerability.

\paragraph{Generative editing.} The emgerging generative editing methods poses a severe challenge to 3DGS IP protection. Diffusion-based 2D editing \cite{brooks2023instructpix2pix,wan2025mft,zhang2023adding} and 3D editing \cite{wu2024gaussctrl,lee2025editsplat,chen2024dge,wang2024view,chen2024gaussianeditor} operate as powerful semantic modifiers. These editing models typically function as aggressive low-pass filters \cite{lu2024robust,zhao2025rdsplat}, filtering out high-frequency protection perturbations in 3DGS assets \cite{zhao2025rdsplat}. Most recent 3DGS watermarking method \cite{zhao2025rdsplat} exploits this observation and explicitly embed watermarks into low-frequency covariance components, ensuring robust 3DGS watermarking against generative editing. Alternatively, 3DGS editing safeguard approaches \cite{hong2025adlift,meng2025degauss} adopt an active defense strategy to disrupt the unauthorized editing process itself.

%% file: sec/7_future.tex
\vspace{-1mm}
\section{Future Directions and Challenges}
\label{sec:future}

Although 3DGS IP protection has made notable progress, numerous unaddressed challenges remain. In this section, we outline six key future directions with their challenges, grouped into three thematic areas: robustness (\textbf{R1-3}), efficiency (\textbf{E1-2}), and protection paradigms (\textbf{P1}).

\paragraph{R1: Resistance to generative purification and editing.} 
Generative 2D purification \cite{nie2022diffusion,xue2024pixel,zhao2024can,cao2023impress} and emerging 3D purification strategies \cite{huang2025can} can effectively eliminate invisible protective signals, whereas most existing methods do not explicitly incorporate purification-aware training or evaluation, resulting in fragile real-world robustness.
Besides, for generative editing \cite{wu2024gaussctrl,lee2025editsplat,chen2024dge,wang2024view,chen2024gaussianeditor},
current defense \cite{hong2025adlift,zhao2025rdsplat} often compromise visual quality to achieve resistance against diffusion-based editing \cite{zhao2025rdsplat}. Pushing the frontier to achieve both high imperceptibility and resilience against generative distortions remains a grand challenge.

\paragraph{R2: The cross-representation survival problem.} In real-world distribution pipelines, 3DGS assets are frequently converted into other formats such as meshes, voxels, or NeRF \cite{mildenhall2021nerf}. Most existing methods fail to survive these representation conversions. While initial attempts like MantleMark \cite{luo2025mantlemark} explore 2D-to-3D survival, future research must explore representation-agnostic perturbations techniques \cite{song2024geometry,huang2025towards}. These should embed information into high-level geometric topology or frequency domains \cite{wang2025stegagaussian}, ensuring the watermark persists even when the 3DGS is converted to other formats \cite{mildenhall2021nerf,liu2025hide,wang2024unlearnable}.

\paragraph{R3: Unified benchmarking.} 
The lack of a standardized evaluation framework hinders fair comparison across methods. Existing works rely on different datasets, attack configurations, and evaluation metrics. As shown in \Cref{tab:survey_3dgs_robustness}, most methods do not consider or evaluate robustness against diverse attack types. To enable systematic progress, the community urgently needs a unified benchmark that provides a standardized attack suite (ranging from conventional distortions to generative attacks), a fixed set of diverse test scenes, and unified metrics for imperceptibility and robustness.

\paragraph{E1: Generalizable and universal perturbations.} 
Most 3DGS IP protection methods \cite{jang20253d,chen2025guardsplat} rely on per-asset fine-tuning, which is computationally expensive and impractical for large-scale asset pipelines. An important future direction is the development of generalizable mappings or universal perturbations \cite{li2025gs,huang2025marksplatter}. Such approaches should be pre-trained on large-scale 3D datasets, enabling perturbations to be injected into arbitrary 3DGS assets in a single forward pass at inference time, without the need for per-asset fine-tuning.

\paragraph{E2: Built-in security for reconstruction and generative 3DGS.} 
As the primary source of 3D assets shifts to feed-forward reconstruction \cite{jiang2025anysplat} and generative pipelines \cite{xiang2025structured}, IP protection must shift accordingly. Future feed-forward reconstruction and large-scale 3DGS generative models should incorporate built-in IP protective capabilities \cite{li2025gaussianseal,li2025instantsplamp}, where perturbation mechanisms are integrated directly into the model's weights or latent space. Consequently, such models would produce 3DGS assets that are inherently protected, potentially eliminating the need for post-hoc protection.

\paragraph{P1: Protection paradigm.}

Existing research on 3DGS IP protection mainly focus on post-hoc verification paradigms such as 3DGS watermarking \cite{huang2024gaussianmarker} and 3DGS steganography \cite{li2024gaussianstego}, which focus on ownership identification rather than usage control. In contrast, active protection, i.e., restricting a protected asset's usability for downstream tasks, remains largely underexplored, with recent progress only beginning to address anti-editing protection against instruction-driven manipulation \cite{hong2025adlift,meng2025degauss}.
However, comprehensive lifecycle asset protection should span the entire pipeline, from initial non-usability to post-training deletability \cite{li2025rethinking}, encompassing mechanisms such as unlearnability \cite{huang2021unlearnable}, non-transferability \cite{wang2021non,hong2025toward,hong2025data,xiang2025jailbreaking,hong2024your,hong2024improving,feng2025token}, and data forgetting \cite{bourtoule2021machine,wang2025machine,yang2025exploring,gao2024large}, among others.
Developing such lifecycle-aware protection mechanisms is foundational for establishing a healthy ecosystem for 3DGS asset IP governance in the AIGC era, enabling technical enforceability, traceability, and legal auditability.

%% file: sec/8_conclusion.tex
\vspace{-1mm}
\section{Conclusion}
\label{sec:conclusion}

In this survey, we provide the first structured and comprehensive overview of 3DGS IP protection. We introduce a bottom-up framework that spans three complementary layers, from underlying Gaussian-based perturbation mechanisms, to passive and active 3DGS IP protection paradigms, and to emerging robustness threats in the AIGC era. 
We also outline key research challenges and future opportunities across robustness, efficiency, and protection paradigms, with the goal of facilitating trustworthy and scalable deployment of 3DGS assets.
We hope this survey serves as a foundational step toward building a trustworthy and secure ecosystem for 3DGS IP applications and enables future research to develop more robust, efficient, and deployable protection solutions.